Hiding in Plain Sight: Towards the Science of Linguistic Steganography[1]

Leela Raj-Sankar[2]
Hamilton High School
Chandler, Arizona, USA
leelars25@gmail.com

S. Raj Rajagopalan
Computer Science Department, Tandon School of Engineering
New York University
Brooklyn, New York, USA
sr6268@nyu.edu

*Extended Abstract*

INTRODUCTION

Steganography is the act of communicating an alternate message inside an otherwise normal-looking communication in a way that the intended communicators can communicate without third parties being able to detect even the presence of the alternate message. This is achieved by having a signal or message called *cover* that looks normal to third parties with an included *covert* message that only the intended interlocutors can understand. The undetectability of the covert message in the cover is the most important feature and requirement of steganographic systems. In comparison, encrypted communications, while granting confidentiality of the hidden message, are easily detected as such. Encrypted communication looks very different from normal unencrypted communication in a variety of ways, and thus cannot meet the undetectability requirement.

A simple example can be found in a kind of magician's show. In this trick, the magician performer on stage volunteers to be blindfolded while his assistant walks among the audience asking for the members to show him something that they have that is hidden (such as the contents of their wallet). Somehow the blindfolded magician is able to guess correctly the various items that the audience has revealed. The audience is dumbfounded but the explanation is simple. The magician's assistant is constantly talking to the audience members throughout the performance and while it sounds like a commonplace conversation to the audience it is secretly signaling to the blindfolded magician who has agreed to a secret code with the assistant. The astounding fact is that even after this fact is revealed to the audience that the assistant is communicating with the magician, the audience is unable to understand how it is happening. Everything the assistant is saying is apparently "normal", he is not using any special signals or exotic words and yet the magician is extracting the requisite information. This is an elementary example of steganography -- the commonplace words or phrases such as "higher", "is that new", "hold it to the light" etc are all designed to sound normal in the context of the magic trick but part of an elaborate code telling the magician what that audience member is holding in their hand. Thus the message from the assistant to the magician is audible to all but the underlying meaning and purpose are only known to the performers and the audience is unable to even sense that this covet communication is going on.

Steganography systems have been discussed and designed for a variety of specific communication media such as images and videos. The design of steganographic systems is specific to the type of communication channel being used for cover and is dependent on the syntax and semantics of the channel. *Linguistic* steganography (LS) is the act of steganography in verbal/spoken communications and is specific to the language and the genre of verbal communication. The practice of LS is very common in certain spoken English domains such as magician performers, "mentalist" shows, etc, however research in systematic systems to build reliable LS systems for commonplace spoken/verbal English is sparse. This project aims to create a program that systematically generates verbal English steganographic codes for and develop a quantifiable mathematical framework to determine effectiveness, based on the results of two primary questions establishing parameters of decodability (probability that the receiver of the coded message will decode the message correctly), density (ratio of codewords to total), and detectability (probability that an arbitrary third party can tell the difference between an untampered cover compared to a *steganized* version of it). Since the smallest semantically meaningful unit of spoken communication is the word, the approach taken is to identify word frequencies within a representative corpus or dataset, insert code words of various frequencies into appropriate positions in the cover text (such as a sentence or a tweet), and validate the number of decoding errors due to extra code words. By definition, a covert message inserted into a cover will have to take the form of a word or groups of words that already appear in cover messages to evade detection. The construction of the covert message insertion can take a variety of forms such as choice of specific words, choice of words containing a specific letter, choice of location of specific words, etc. The insertion of code words into highly probable cover text distorts the word frequencies and may create semantically incorrect or unusual text, which could lead to detection. For concreteness and ease of understanding, we analyze a scheme of inserting specifically chosen codewords into cover text, however the approach followed should work for any method specifically selecting words in a cover. The focus of this research is on creating a system of inserting code words into the cover text in such a way that the word frequency distortions and likelihood of semantic errors is minimized.

---
[1] This is an extended description of a poster of the same title that was presented at the 2020 Information Theory and Applications Workshop held in San Diego, CA.
[2] Presenting author

UNIQUE CHALLENGES IN STEGANOGRAPHY
There are no known methods of measuring semantic distortion in spoken in an entirely *local* manner, i.e. methods to measure semantic distortion without looking at the entire corpus or a significant portion of it for every insertion decision. In natural language, context is an important feature of semantic correctness. However, it is difficult to establish boundaries of semantic correctness -- whether a word or phrase is likely to be used in a context may be correlated with its immediate context (such as the conversation or email or news article) or its larger context (such as historical references). Locality is an essential feature of a *practical* LS system since it is meant to be used easily by a program or a speaker alike, who need a quick and easy method of using the system without having constant access to the dataset. The key innovation in this research is to use n-gram word frequencies as a measure of *local* semantic meaningfulness. This project created n-gram distributions of a given corpus or data set which guides the insertion of code words to minimize distortion of n-gram frequencies. N-grams of words correspond to phrases in English, and so the approach focuses on inserting codewords into the right phrases that would naturally appear in a cover message belonging to that particular genre.

BASICS OF STEGANOGRAPHY
The basics of a steganography scheme are illustrated in the following figure. Alice wants to communicate the message "43" to Bob on a public channel. If Alice were to encrypt her message, it would look like the bottom of the figure. However, it suffers from the fact that it is obvious that Alice is trying to send *some* information. What if Alice wants to hide the very fact that she is communicating something secret to Bob? In the proposed scheme, Alice and Bob have agreed a priori to a code (represented by the "A") that maps in some unspecified manner, for example, "take" to "4" and "pill" to "3" (or alternately, the combination of "take" and "pill" to "43"). An eavesdropper must not be able to tell that Alice is communicating anything other than what is being broadcast. This means, among other things, that the message "Take the red pill" from Alice to Bob should look like a normal communication from Alice to Bob given the entire context which in this case is the historical transcript of communications between Alice and Bob. It is the undetectability feature that makes steganography a greater challenge than confidentiality.

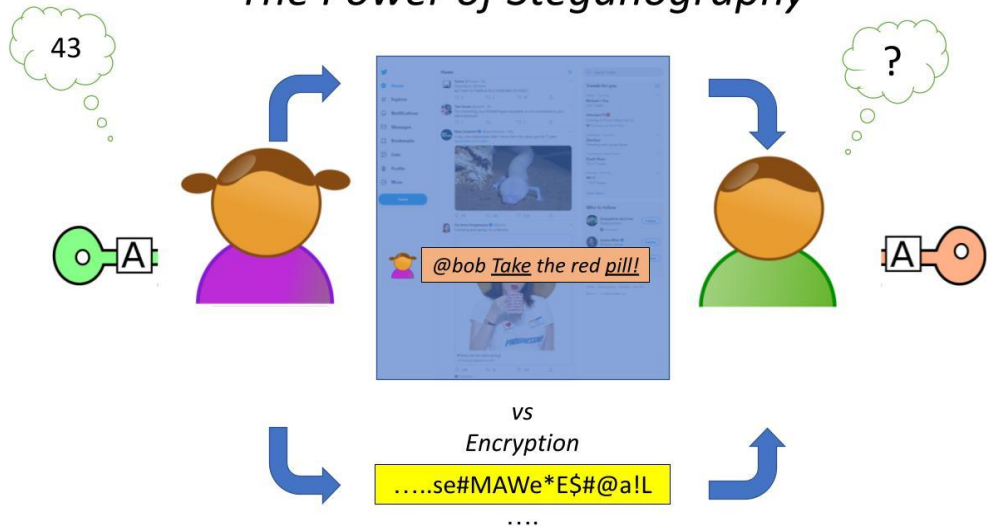

CHALLENGES IN BUILDING A PRACTICAL STEGANOGRAPHY SCHEME

Creating a practical LS scheme comes with its own challenge -- the need for a representative data set of verbal/spoken English that can be used to demonstrate and validate an approach. As in any natural language, there are many genres of verbal English ranging from formal language (as used in this paper) to legal communications, song lyrics, Shakespearean drama, etc. For this research project, the approach was to choose a genre that is closest to the spoken English language within a well-defined domain, with the additional requirement of an available representative data set. For this research project a well-known publicly available tweet database was chosen. Tweets are well-known to be, among all written forms of common use language, to be close to the spoken form. The Twitter data set also gives us a well-structured domain within which word frequencies can be measured and compared for distortion. To keep with the motivation for a verbal LS scheme, the use of specific constructs that are available in printed text such as punctuation marks, strategic use of spaces between words, font changes, paging, etc are all eschewed in the construction of the proposed LS scheme. Finally, tweets are a *broadcast medium,* in which anyone who cares can find and read a tweet, and thus a successful steganization scheme for tweets would require the highest level of undetectability..

The minimization of n-gram frequency distortion creates a tradeoff with the density of codeword insertion. As the number of covert symbols inserted into the cover increases, the number of code words inserted increases as well, resulting in increased frequency distortion of the n-grams in the steganized cover, which degrades the undetectability of the code. Minimizing frequency distortions also creates a tradeoff between code world selection and decodability. The codewords that would likely lead to minimal frequency distortions would be the most likely words or n-grams in the data set such as "is" or "the". However, choosing high frequency words as code words runs into the problem of false codes, i.e. a receiver may mistakenly decode a word that was not part of the covert message. Therefore there is a tradeoff between the popularity of a code word and its decodability (the common word case). At the same time, the use of a relatively rare word as a code word will increase the likelihood that the steganization is detected. which causes the choice of code words that have middle-ground frequency in the data set. Thus, there is a robust requirement for choosing code words that are too common (which would degrade decodability) but also not too rare (which would degrade the detectability).

# A MATHEMATICAL MODEL FOR STEGANOGRAPHIC SCHEMES

A mathematical formalism for capturing the approach and the model is presented below. First, the basic definitions of a steganizing scheme.

- Let $\mathfrak{C}$ be the specific universe of cover messages (text from a given domain e.g. Tweets, email, song lyrics) over its word set $W$ and S be a set of codewords or phrases formed from $W$.
- A **Covert Code** is a mapping $\mathfrak{C} \to \mathfrak{C}$ as follows:

$$z = c \pm α(s)$$

    where $c \in \mathfrak{C}$ represents the unmodified original **cover message**, $\pm$ represents **code word insertion process** (insertion could also be substituted for by any one-to-one or many-to-one replacement), $α$ represents code word **mapping** (of secret to codeword from S), **s** represents the **secret message**, and $z \in \mathfrak{C}$ represents the **covert code.**

- A legitimate receiver **R** performs the following transformation over $\mathfrak{C}$:

$$s = α^{-1}(z \mp c)$$

    where $\mp$ is the operation of extracting the codeword(s) from the covert code **z** and $α^{-1}$ represents the inverse mapping of α from codeword to secret message.

Next, the definitions of the concepts of decodability, density, and detectability are presented..
- **Decodability** (probability that the receiver of the coded message will decode the cover correctly), can be defined as:
    - $D = \text{Prob} ( \text{Dec}_R(z) = s)$
    
    where $\text{Dec}_R(.)$ is the decoding algorithm of a legitimate receiver and the probability is calculated over all choices (including random) available to the receiver
- **Density** (ratio of code words to total words in the cover message) can be defined as:
    - d = # of code words / # of total words in the covert code
- **Detectability** (probability that an attacker can tell the difference between an untampered cover compared to a steganized version of it) can be defined as
    - $\Delta = \text{Prob} (\text{Dec}_A(z,c) \text{ is correct})$
    - Where $\text{Dec}_A(.)$ is a boolean algorithm for any non-legitimate receiver A that is able to distinguish cover from covert code, and the probability is over all choices (including random) available to A.
- The **ideal** covert code is one which has $D \to 1, d \to 1, \Delta \to 0$
- **D** and **d** can be measured definitely but $\Delta$ can only be measured empirically or proven mathematically.

Finally, the mathematical model for the security of the steganographic scheme is given below. For this model, we borrow terms from Cryptographic literature. Note that the inverse polynomial limit for undetectability is used here merely to complete the formalism. In practice, we need empirical limits on undetectability for finite sized cover sets.

- To **completely break the code**, a non-legitimate receiver A has to perform the transformation **s'** = $β$**(z)** where $β$(.) is any algorithm of A's choice given all the public information. A successful break is when **s' = s** happens with high probability.
- **To make the definition of covertness stronger**, a non-legitimate receiver A only has to **consistently distinguish** cover text from covert code,
- i.e. given any pair (c,z), the covert code is not detectable if the **probability that A can**

- **consistently tell which is which is almost or equal to zero**. *This definition underscores the difference between encryption and steganography because A does not need* to extract s to be successful.
- Detectability of the code may be affected by **the size n of the cover set ℭ**
- Inspired by the definition of security for encryption algorithms, we propose this definition of undetectability.
    - Given a cover set ℭ with n elements, for all algorithms A, a covert code is **undetectable** if

        $$\Delta_n = \text{Prob}(\textbf{Dec}_A(z,c) \text{ is correct}) < 1/n^{o(1)}$$

    - In plain English, the probability of detection decreases faster as a function of n than any inverse polynomial. Ideally

For an arbitrary attacker, the only way to guarantee that the steganization is undetectable is using information theoretic considerations. In other words, the steganized covers can only be indistinguishable from unsteganized ones is if the probability distributions are the same. More formally, this requires that the statistical distance, also known as the Kullback-Leibler divergence $D\_KL(P||Q)$ be small where P is the probability distribution of cover messages and Q is the distribution of steganized cover messages. However, KL distance is a very compute-intensive task even for small datasets and is thus not a viable method for computing the codeword that minimizes the distance. Our proposal for using n-grams to minimize this distortion can be considered an approximation of the statistical distance that is relatively easy to compute.

A PRACTICAL CONSTRUCTION OF STEGANOGRAPHY FOR TWEETS

For the practical demonstration of the concepts here the research used a corpus of approximately 240,300 Tweets from the Stanford Sentiment Analysis Dataset. After scrubbing the dataset of personal information (such as user names) and characters that cannot be used in spoken language (such as hashtags), the word frequencies were computed using the Python Natural Language Toolkit (nltk). First ten relatively common codewords were randomly chosen to represent the ten digits 0-9. This secret mapping is known only to the sender and receiver.

To demonstrate the steganized transmission of a single random digit, the Program inserted one or more code words at a time into a randomly chosen Tweet from the data set. The next step is to check if the steganized Tweet has any accidental code words; if so, the program rejects that Tweet and repeats the process with another randomly chosen Tweet until there are no accidental code words. Before transmitting the steganized cover the program validates that the cover can be decoded correctly.

In the adjoining figure, the random secret message is the number "21". The randomly selected tweet for the cover (extracted from the data set) is "Poor cast off to the trash help when no longer useful". (note that the spelling errors and other idiosyncrasies of the original data are retained). The codewords corresponding to the digits "2' and "1" are "good" and "really". These codewords are sequentially inserted into inter-word positions in the cover. Each insertion position causes a different n-gram to be formed in the tweet. Each of the n-grams is analyzed to see if it is a likely n-gram in the original dataset. This is done for n=1, 2, 3, 4…. As n increases the frequencies of occurrence in the data set change and for large enough n every n-gram becomes unique. Thus the maximal value of n is determined by the data set.

```
Secret message: 2 1
Tweet picked: Poor cast off to the trash heap when no longer usefull

Stego:  Poor cast off to the good trash heap when no longer really usefull
```

This adjoining graph shows the tradeoff between choosing highly common codewords and their effect on decodability. As can be seen, the number of errors in decoding (defined as the receiver extracting anything different from the sent message) increases dramatically as the word frequency increases, thus showing that highly common words are poor candidates for codewords.

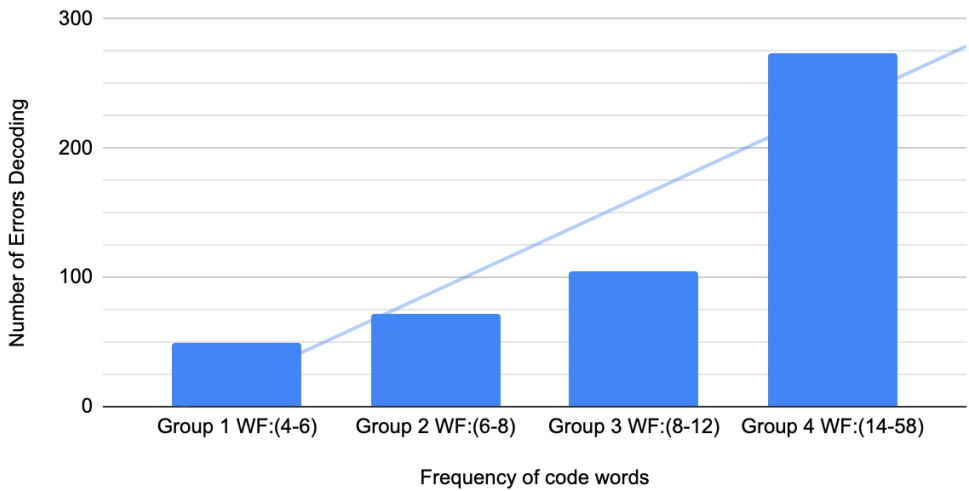

The table below shows the word frequencies (WF) in the original data set, x-y indicates codewords appearing x-y times in the original cover. The errors show many times the steganized tweet that had been decoded incorrectly as a result of choosing highly popular codewords.

|  | Group 1 - WF: 4-6 | Group 2 - WF: 6-8 | Group 3 - WF: 8-12 | Group 4 - WF: 14+ |
|---|---|---|---|---|
| Total # of Errors | 49 | 72 | 105 | 274 |

Next, the effect of the code words on the detectability of the steganization was measured by repeating the experiment with multiple choices of random secret messages and thecover tweet. For various code densities, the word frequencies of the steganized tweets with the frequencies of the original data set were compared. .

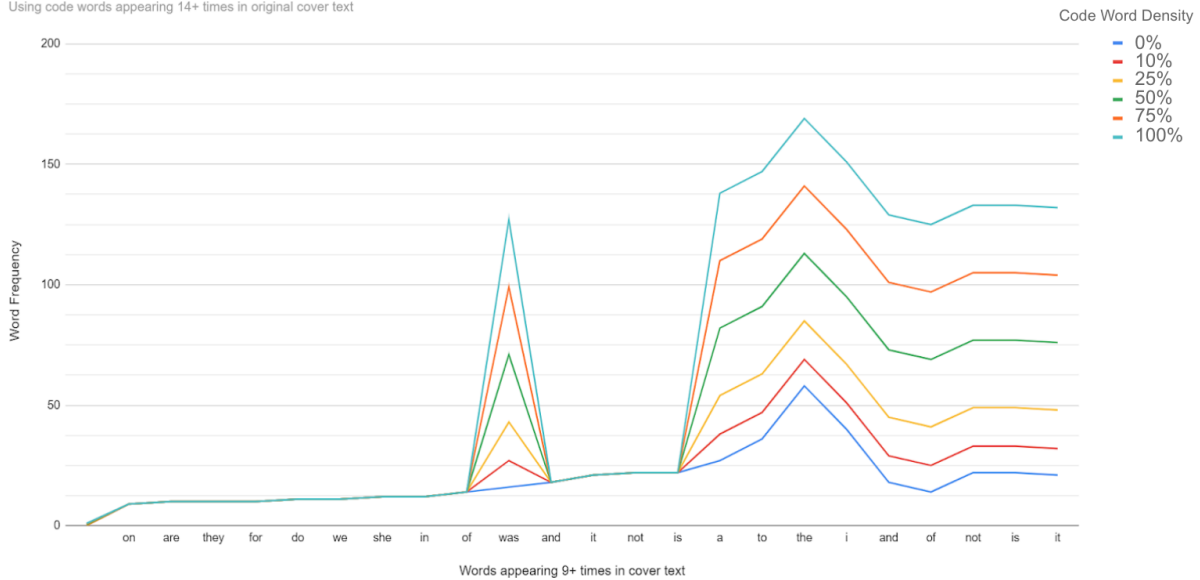

EXTENSIONS AND FUTURE WORK

The steganizing scheme outlined above is applicable to a number of similar scenarios. For example, the scheme easily generalizes to the use of code phrases (or code sentences) instead of codewords. The concept of n-grams applies to these different granularities as well, if at a higher computational cost. Similarly, the steganizing scheme can be used in slightly different mapping schemes. For example, the mapping of the digits can be done probabilistically instead of deterministically. Our formalisms will work exactly as before with some appropriate degradation in the decodability and/or detectability of the code. Similarly, if the mapping of a secret symbol is to the beginning letter of a word alone rather than to an entire word -- say, the digiti 1 is coded with any word beginning with the letter "F" and so on -- the steganographic scheme outlined can be adapted by selecting an appropriate word or phrase that satisfies the constraints. However, note that any such steganization will cause some distortion of the word frequencies of the cover set even if it is less than the distortion caused by the scheme above. The frequency distortions of this scheme can also be handled using an analogous scheme as ours where the choice of the inserted codeword can be done to minimize the n-gram frequency distortion.

One limitation of such a scheme, as in any practical cryptographic scheme as well, is that we have to make some assumptions about what an adversary can or cannot do. As mentioned earlier, our n-gram based scheme for minimizing distortion is only an approximation of the true statistical distortion and is thus susceptible to detection in some unknown way. It is possible that the adversary can detect statistical distortion in some higher dimension than what was used in the code (such as (n+1)-grams). However in practice this is a computational tradeoff -- computing n-grams for large n is computationally expensive, but one can in principle use n large enough such that the cost of computing the distortions becomes computationally prohibitive for the adversary.

In the future, we plan to explore extremely high dimensional codes such as ones produced by neural networks. However, we feel that modeling an adversary for such a steganographic scheme will be equally challenging.